\documentclass{article}

\usepackage[nonatbib, preprint]{neurips_2024}

\usepackage[utf8]{inputenc} 
\usepackage[T1]{fontenc}    
\usepackage{hyperref}       
\usepackage{url}            
\usepackage{booktabs}       
\usepackage{amsfonts}       
\usepackage{nicefrac}       
\usepackage{microtype}      
\usepackage{subfigure}

\usepackage{epsfig}
\usepackage{graphicx}
\usepackage{amsmath}
\usepackage{amssymb}
\usepackage{amsthm}
\usepackage{adjustbox}
\usepackage{caption}
\usepackage{verbatim}

\usepackage{algorithm}
\usepackage{algorithmic}
\usepackage{enumitem}

\usepackage{hyperref}
\usepackage{url}

\usepackage{glossaries}

\newacronym{ssl}{SSL}{Self-Supervised Learning}
\newacronym{mae}{MAE}{Masked Auto Encoders}
\newacronym{mim}{MIM}{Masked Image Modeling}
\newacronym{cnn}{CNN}{Convolution Neural Network}
\newacronym{gnn}{GNN}{Graph Neural Network}
\newacronym{sota}{SOTA}{State-Of-The-Art}
\newacronym{GM3D}{GM3D}{Selective Attention in 3D}
\newacronym{geomask3d}{GM3D}{GeoMask3D}
\newacronym{fps}{FPS}{Farthest Point Sampling}
\newacronym{knn}{KNN}{K-Nearest Neighbours}
\newacronym{ip}{IP}{Informative Patches}
\newacronym{drc}{DRC}{Dense Relation Comparison}
\newacronym{mse}{MSE}{Mean Square Error}
\newacronym{svm}{SVM}{Support Vector Machine}
\newacronym{gc}{GC}{Geometric Complexity}

\usepackage{shuffle}
\usepackage{mathtools}
\usepackage{bbm}
\usepackage{rotating}
\usepackage{comment}
\usepackage{stmaryrd}
\usepackage{cleveref}
\usepackage{amssymb}
\usepackage{colortbl}
\usepackage{multirow}
\usepackage{graphicx}
\usepackage{caption}

\newsavebox\CBox
\def\textBF#1{\sbox\CBox{#1}\resizebox{\wd\CBox}{\ht\CBox}{\textbf{#1}}}

\usepackage{booktabs}
\usepackage{makecell}

\usepackage{stmaryrd}
\usepackage{algorithm}
\newenvironment{indentedblock}{%
  \begin{quote}%
}{%
  \end{quote}%
}

\usepackage[dvipsnames]{xcolor}

\usepackage{wrapfig,lipsum,booktabs}

\newcommand{\mr}[1]{\mathit{#1}}

\newcommand{\selfsup}{\begin{turn}{90}  \begin{tiny}SS\end{tiny}\end{turn}\,}

\newcommand{\Real}{\mathbb{R}}

\newcommand{\Loss}{\mathcal{L}}

\newcommand{\encoder}{\mathcal{E}}
\newcommand{\decoder}{\mathcal{D}}
\newcommand{\vOne}{\mathbbm{1}}

\newcommand{\symbVis}{v}
\newcommand{\symbMasked}{m}

\newcommand{\InputAll}{X}

\newcommand{\RecAll}{\hat{X}}
\newcommand{\LatentAll}{Z}

\newcommand{\InputVis}{X^{\symbVis}}

\newcommand{\LatentVis}{Z^{\symbVis}}
\newcommand{\NumVis}{N^{\symbVis}}

\newcommand{\InputMasked}{X^{\symbMasked}}
\newcommand{\inputMasked}{x^{\symbMasked}}
\newcommand{\RecMasked}{\hat{X}^{\symbMasked}}
\newcommand{\recMasked}{\hat{x}^{\symbMasked}}
\newcommand{\NumMasked}{N^{\symbMasked}}

\usepackage[numbers]{natbib}

\newcommand{\mypar}[1]{\vspace{2.5pt}
\noindent\textbf{#1~}}

\newcommand{\ppm}{\,\scriptsize$\pm$}

\newcommand{\citesmall}[1]{\scriptsize\cite{#1}}

\title{GeoMask3D: Geometrically Informed Mask Selection for Self-Supervised Point Cloud Learning in 3D}

\author{Ali Bahri\thanks{Correspondence to \href{mailto:mehrdad.noori.1@ens.etsmtl.ca}{ali.bahri.1@ens.etsmtl.ca}} \And Moslem Yazdanpanah \And Mehrdad Noori \And Milad Cheraghalikhani \And Gustavo adolfo.vargas-hakim \And David Osowiech \AND Farzad Beizaee \And Ismail Ben Ayed\And Christian Desrosiers\AND \
LIVIA, ÉTS Montréal, Canada \\
International Laboratory on Learning Systems (ILLS), \\
McGILL - ETS - MILA - CNRS - Université Paris-Saclay - CentraleSupélec, Canada}

\def\textBF#1{\sbox\CBox{#1}\resizebox{\wd\CBox}{\ht\CBox}{\textbf{#1}}}

\begin{document}

\maketitle

\begin{abstract}
We introduce a novel approach to self-supervised learning for point clouds, employing a geometrically informed mask selection strategy called \gls*{geomask3d} to boost the efficiency of \gls*{mae}. Unlike the conventional method of random masking, our technique utilizes a teacher-student model to focus on intricate areas within the data, guiding the model's focus toward regions with higher geometric complexity. This strategy is grounded in the hypothesis that concentrating on harder patches yields a more robust feature representation, as evidenced by the improved performance on downstream tasks. 
Our method also presents a 
feature-level knowledge distillation technique designed to guide the prediction of geometric complexity, which utilizes a comprehensive context from feature-level information.
Extensive experiments confirm our method's superiority over \gls*{sota} baselines, demonstrating marked improvements in classification, segmentation, and few-shot tasks. Code is available on \href{https://github.com/AliBahri94/GM3D.git}{\emph{our GitHub}}.
\end{abstract}

\section{Introduction}
\label{sec:intro}

\begin{figure}
    \centering
    \includegraphics[width= 0.5 \linewidth]{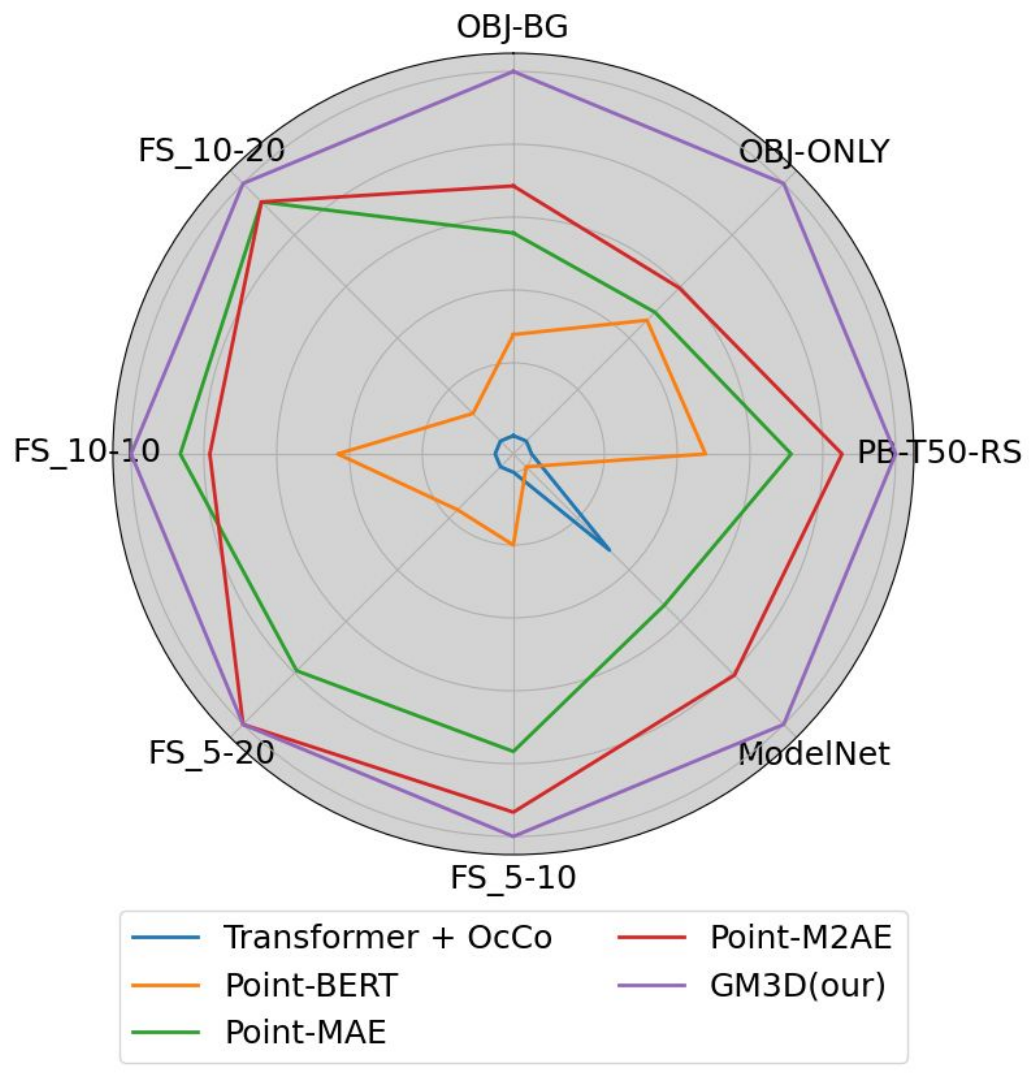}
    \caption{\label{fig:radar} A relative comparison of the \gls*{sota} point cloud MAE methods on different tasks. Here, the center and the outer circles represent the lowest and highest values on each task, respectively.}
\end{figure}

The advent of large-scale 3D datasets~\cite{deitke2023objaverse, slim2023_3dcompatplus} has propelled research on deep learning for point clouds, leading to notable improvements in complex 3D tasks. However, the time-consuming nature of data collection, compounded by the complexity of 3D view variations and the mismatch between human perception and point cloud representation, significantly hinders the development of effective deep networks for this type of data \cite{xiao2023unsupervised}. In response to this challenge, \gls*{ssl} has emerged as a promising solution, facilitating the learning of representations without relying on manual annotations. \gls*{ssl} not only circumvents the issues of costly and error-prone labeling but also improves the model's generalization ability, offering a pivotal advancement in the field of point cloud-based deep learning \cite{fei2023self}.

\gls*{mae}s, as simple yet effective self-supervised learners, have gained prominence by learning to recover masked parts of data. This approach has significantly advanced NLP models and has resulted in exceptional vision-based representation learners \cite{kenton2019bert,he2022masked} when applied to vision tasks. Building on their success, \gls*{mae}s have recently been adapted for point cloud representation learning, leading to \gls*{sota} methods including MaskPoint \cite{liu2022masked}, Point-Bert \cite{yu2022point}, Point-MAE \cite{pang2022masked}, Point-M2AE \cite{zhang2022point}, I2P-MAE \cite{zhang2023learning}, and MAE3D \cite{jiang2023masked}.
However, these methods share a common limitation stemming from the random masking strategy they use, where masked regions of the point cloud are selected arbitrarily without taking into account their informativeness. As demonstrated in recent work on \gls*{mim}~\cite{wang2023hard}, employing a selection strategy that prioritizes informative regions over background areas can significantly enhance the robustness of the learned representation. While such a strategy has shown promising results for image processing, its application to point clouds has not been explored so far.

\begin{figure*}[t]
  \centering
   \includegraphics[width=.9\linewidth]{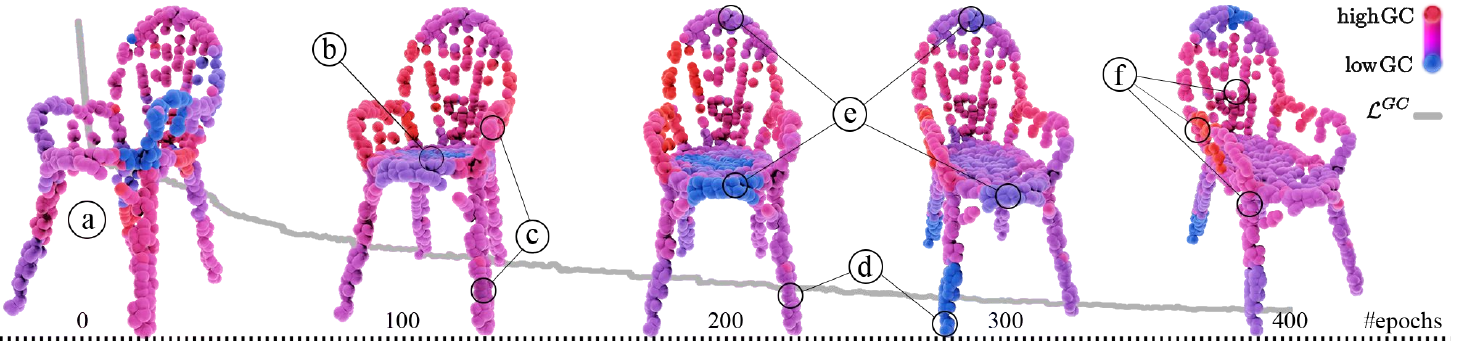}
   \caption{
   \small Visualization of estimated \gls*{gc} progression throughout training is depicted. The color spectrum denotes \gls*{gc}, ranging from low (\textcolor{blue}{Blue}) to high (\textcolor{red}{Red}). \gls*{gc} values are normalized per object to reflect relative complexity across patches within each object's point cloud. As training progresses (from left to right), initial \gls*{gc} rankings display a random distribution (a). After 100 epochs, the model learns to assign lower complexity rankings to smooth areas (b) and higher rankings to complex regions (c). Through \gls*{gc} guided masking, the model increasingly focuses on complex areas from epochs 200 to 300, resulting in a reduction of \gls*{gc} ranking (d) and smoothing of the complexity ranking distribution, accompanied by a decrease in total complexity loss $\Loss^{GC}$ (e). Eventually, the model converges to a low $\Loss^{GC}$ value, consistently targeting canonical patches while maintaining a smoother \gls*{gc} distribution (f).
   }
   \label{fig:banner}
\end{figure*}

To bridge this gap, we study the use of a targeted masking strategy for point clouds within the MAE framework, applicable across both single and multi-scale methods. 
We introduce \acrfull*{geomask3d}, a novel geometrically-informed mask selection strategy for object point clouds. Due to the lack of background data in the object's point clouds, \gls*{geomask3d} enables models to concentrate on more complex areas, such as canonical ones with higher connections to the other areas, and pay less attention to the geometrically simple areas like smooth surfaces, as depicted in Fig.~\ref{fig:banner}. 
To showcase the effectiveness of our method, we integrate it into the pretraining process of both single-scale Point-MAE and multi-scale Point-M2AE, the leading \gls*{mae} methods for point clouds. As illustrated in Fig.~\ref{fig:radar}, our method exhibits notable enhancements over the earlier \gls*{sota} approaches across a range of challenging tasks.

To the best of our knowledge, this represents the first attempt to implement a masking strategy for point clouds, independent of additional modalities such as multi-view images. Our contributions are summarized as follows:
\begin{enumerate}\setlength\itemsep{0em}
    \item We propose a novel masking approach for point cloud \gls*{mae}s, which selects patches based on their geometric complexity rather than selecting them randomly. This approach employs an easy-to-hard curriculum learning strategy where the ratio of patches selected using geometric complexity is gradually increased during training.
    \item We also introduce a 
    feature-level knowledge distillation technique to further guide the prediction of geometric complexity. Instead of relying on the noisy and incomplete information of 3D points, this efficient technique transfers latent features from a frozen teacher model, encoding higher-level information on the geometry, to the student model learning the point cloud representation.  
    
     

    \item We integrate these mechanisms into the pretraining process of single-scale Point-MAE and multi-scale PointM2AE, both of which are \gls*{sota} point cloud MAEs, significantly enhancing their performance in diverse downstream tasks. 
\end{enumerate}

\vspace{-10 pt}

\section{Related Works}

\mypar{Point Cloud Learning.} PointNet \citep{qi2017pointnet} established point cloud processing as a key method in 3D geometric data analysis by addressing the permutation issue of point clouds with a max-pooling layer. 
To further enhance performance and capture both local and global features, PointNet++ \citep{qi2017pointnet++} introduced a hierarchical structure, expanding the receptive fields of its kernels recursively for improved results over PointNet. Another study \citep{jaritz2019multi} focused on point cloud scene processing, where multi-view image features are combined with point clouds. In this study, 2D image features are aggregated into 3D point clouds and a point-based network fuses these features in 3D space for semantic labeling, demonstrating the substantial benefits of multi-view information in point cloud analysis.

\mypar{\gls*{mae} for Representation Learning.} Leveraging the success of \gls*{mae}s in text and image modalities, Point-BERT \citep{yu2022point} introduced an approach inspired by BERT \citep{devlin2018bert} adapting Transformers to 3D point clouds. This approach creates a Masked Point Modeling task, partitioning point clouds into patches and using a Tokenizer with a discrete Variational AutoEncoder (dVAE) to produce localized point tokens, with the goal of recovering original tokens at masked points. Similarly, Point-GPT~\citep{chen2024pointgpt} introduced an auto-regressive generative pretraining (GPT) approach to address the unordered nature and low information density of point clouds. ACT~\citep{dong2022autoencoders} proposed a cross-modal knowledge transfer method using pretrained 2D or natural language Transformers as teachers for 3D representation learning. MaskPoint \citep{liu2022masked}, a discriminative mask pretraining framework for point clouds, represents the point cloud with discrete occupancy values and performs binary classification between object points and noise points, showing resilience to point sampling variance. Point-MAE \citep{pang2022masked} adapted \gls*{mae}-style pretraining to 3D point clouds, employing a specialized Transformer-based autoencoder to reconstruct masked irregular patches and demonstrating strong generalization in various tasks. Following this, MAE3D \citep{jiang2023masked} used a Patch Embedding Module for feature extraction from unmasked patches. Point-M2AE \citep{zhang2022point} introduced a Multi-scale \gls*{mae} framework with a pyramid architecture for self-supervised learning, focusing on fine-grained and high-level semantics. I2P-MAE \citep{zhang2023learning} further improved the self-supervised point cloud learning process by leveraging pretrained 2D models through an Image-to-Point transformation. 

Our proposed method, which can be integrated into any point cloud \gls*{mae} architecture, differs from previous approaches like Point-MaskPoint, MAE, MAE3D and Point-M2AE that are based on random patch selection. Moreover, unlike recent point cloud learning approaches such as I2P-MAE \citep{zhang2023learning}, which rely on image information and 2D backbones, our method only requires 3D coordinates as input.

\section{Method}
\label{sec:method}

\captionsetup{belowskip=-10pt}

\subsection{Preliminaries}

\mypar{Masked Auto Encoders.} Autoencoders use an encoder $\encoder$ to map an input $\InputAll$ to a latent representation $\LatentAll = \encoder(X)$ and a decoder to reconstruct the input as $\RecAll = \decoder(\LatentAll)$. \acrfull*{mae} are a special type of autoencoder that receive a masked-patch input composed of a set of visible patches (with positional encoding) $\InputVis$ and the index set of masked patches $M$, and reconstruct the masked patches as follows: 
\begin{equation}
    [\InputVis, \RecMasked] = \mathrm{MAE}(\InputVis, M) = \decoder\big([\encoder(\InputVis), T_M]\big)
\end{equation}

The encoder $\encoder$ and decoder $\decoder$ are both transformer-based networks. The encoder only transforms the visible patches to their latent representations $\LatentVis = \encoder(\InputVis)$. On the other hand, the decoder takes as input $\LatentVis$ and a set of tokens $T_M = \big\{\big(t_{\mr{mask}}, E_{\mr{pos}}(i)\big) \, | \, i \in M\big\}$ where $t_{\mr{mask}}$ is a global learnable mask token and $E_{\mr{pos}}(i)$ is the positional embedding of masked patch $i\in M$. Let $N = \NumVis + \NumMasked$ be the number of patches in the input, with $\NumVis=|\InputVis|$ and $\NumMasked=|M|$, the masking ratio is defined as $m^{\mr{ratio}} = \NumMasked/N$.

 \begin{figure*}[t]
  \centering
   \includegraphics[width=0.9\linewidth, height=6.5cm]{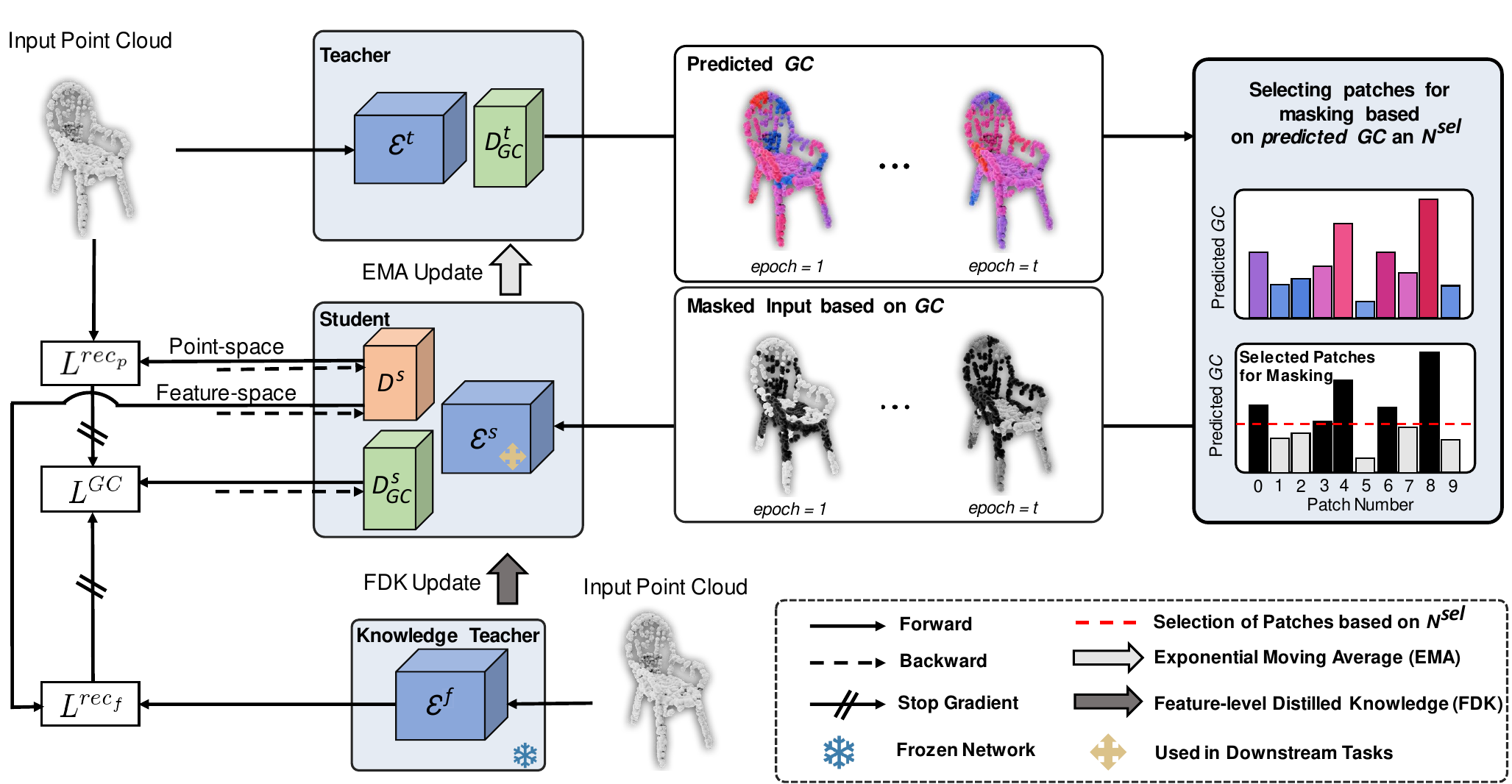}
   \caption{Overview of the \acrfull*{geomask3d} method for self-supervised representation learning in point clouds. The Teacher network predicts \acrfull*{gc}, and patches with the highest \gls*{gc}, denoted by $N^{\mr{sel}}$, are selected for masking. The Student network is then trained to reconstruct these masked tokens while simultaneously learning \gls*{gc} through the loss $\Loss^{GC}$. The reconstruction loss is defined as $\Loss^{\mr{rec}} = \Loss^{\mr{rec}_p} + \Loss^{\mr{rec}_f}$. The Teacher network's weights are updated using the Exponential Moving Average (EMA) of the Student's weights, while the Knowledge Teacher remains frozen and is used for generating encoder features essential for the Student's training with $\Loss^{\mr{rec}_f}$.
}
   \label{fig:diagram}
\end{figure*}

\mypar{\gls*{mae} for Point Clouds.} A 3D point cloud $P$ is a set of $N^p$ points $p_j \in \Real^3$. For this type of data, patches correspond to possibly overlapping subsets of $K$ points in $P$. While there are various ways to generate patches from a point cloud, we follow the strategy employed by several related methods \citep{pang2022masked,zhang2022point} where each patch $x_i$ is defined as a center point $c_i$ and the set $P_i \subset P$ of \gls*{knn} to this center. To uniformly represent the whole point cloud, patch centers $c_i$ are obtained using a \gls*{fps} algorithm, where a first center is randomly chosen from $P$ and then the next one is selected as the point in $P$ furthest away from previously selected centers. Assuming that $c_i$ is included as the first element of its nearest-neighbor list $P_i$, we can represent the patchified version of the point cloud as a tensor $\InputAll \in \Real^{N \times K \times 3}$.

In autoencoders for images, a pixel-wise L2 reconstruction loss is typically used for training the MAE. In our case, since patches $x_i \in X$ are points sets, we instead employ the Chamfer distance to measure the reconstruction error $\Loss^{\mr{rec}_p}$ of masked patches:
\begin{equation}
    \Loss^{\mr{rec}_p} \, = \, \frac{1}{N^m} \sum_{i=1}^{N^m} \mathrm{Chamfer}(\inputMasked_i, \,\recMasked_{i}),
\end{equation}
where the Chamfer distance between two sets of points $S$ and $S'$ is defined as
\begin{equation}
 \mathrm{Chamfer}(S,S') \, = \, 
 \sum_{p \in S} \, \min_{p' \in S'} \| p - p'\|_2^2 \  + \, \sum_{p' \in S'} \, \min_{p \in S} \| p - p'\|_2^2.
\end{equation}
In the next section, we build on these definitions and present our \acrlong*{geomask3d} method for self-supervised point cloud representation learning.

\subsection{\acrlong*{geomask3d}}
In the originally proposed \gls*{mae}, masked patches are selected randomly during each iteration without considering the varying impacts that different patches may have on the training process. This random masking approach may not be efficient, as patches in a point cloud can exhibit varying levels of \acrfull*{gc}, which pose different degrees of challenge to the learning network. Inspired by the principles of human learning—where repeatedly tackling challenging tasks enhances performance over time—we propose prioritizing geometrically complex patches during the pre-training phase of a \gls*{mae} network. A \gls*{mae} network can achieve more efficient and effective learning by shifting from random masking to a strategy focusing on complex patches.

This strategy raises a fundamental question: what is \acrfull*{gc}, and how can it be measured? We define \gls*{gc} for a patch as the relative difficulty of reconstructing that patch using an \gls*{mae} network.  Specifically, a patch is considered complex if the \gls*{mae} network demonstrates difficulty in reconstructing it, as indicated by higher reconstruction loss $ \Loss^{\mr{rec}_p}$.

To this end, we propose \acrfull*{geomask3d} as a modular component that integrates with any point cloud \gls*{mae} backbone. This component is incorporated into the pretraining phase of a chosen method, shifting from a basic naive random masking approach to a selective focus on geometrically complex patches for masking. The architecture of \gls*{geomask3d} employs an auxiliary head $\decoder_{GC}$ for predicting the geometric complexity $GC \in \Real^{N}$ across the input patches, which is trained with a loss $\Loss^{GC}$.

A teacher-student framework is utilized
to integrate \gls*{geomask3d} with a target network. We denote as $\mathrm{\gls*{geomask3d}}^s\!=\!(\encoder^s, \decoder^s, \decoder_{GC}^s)$ the student and as $\mathrm{\gls*{geomask3d}}^t\!=\!(\encoder^t, \decoder^t, \decoder_{GC}^t)$ the teacher, both of them having their own encoder, decoder and auxiliary head.  
In line with \citep{he2020momentum}, we apply a momentum update method to maintain a consistent teacher, updating it in each iteration,
\begin{equation}
    \mathrm{GM3D}^t \, = \, \mu\!\cdot\! \mathrm{GM3D}^t \, +\, (1\!-\!\mu)\! \cdot\!\mathrm{GM3D}^s
    \label{eq:momentum}
\end{equation}
where $\mu$ represents the momentum coefficient. Both networks predict the \gls*{gc} based on the patch's informational content, as elaborated in Section~\ref{sec: Prediction of Geometric Complexity}. We employ the prediction of \gls*{gc} in the masking strategy of the method during its pretraining stage. The \gls*{gc} of the student network (\gls*{gc}\textsuperscript{s}) is predicted based on the masked input $\InputVis$, while the \gls*{gc} of the teacher (\gls*{gc}\textsuperscript{t}) is calculated in inference mode using the complete input $\InputAll$:
\begin{equation}
    GC^a = 
    \begin{cases} 
        \decoder_{GC}^a(\encoder(\InputAll)) ,& \text{if } a = t \ \ \mr{(teacher)}\\
        \decoder_{GC}^a\big([\encoder(\InputVis), T_M]\big) ,& \text{if } a = s \ \ \mr{(student)}
    \end{cases} 
\end{equation}

The overview of our method for self-supervised representation learning in the point cloud is depicted in Fig.~\ref{fig:diagram}.
The \acrfull*{geomask3d} approach involves three interconnected steps, which will be explained in the following sections. Additionally, we provide a detailed explanation of the Knowledge-Distillation-Guided GC strategy in Section~\ref{sec: Knowledge Distillation-Guided}.


\subsubsection{Prediction of \acrfull*{gc}}
\label{sec: Prediction of Geometric Complexity} 

During this stage, our goal is to evaluate \gls*{gc} of each patch in $\InputMasked$, relative to the others within the same set. We achieve this by using a \gls*{drc} loss \citep{wang2023hard} which enforces the \gls*{gc} of masked patch pairs $(k,l)$, predicted by the student (i.e., $GC_{k}^s$ and $GC_{l}^s$), to follow the same relative order as their loss values $\Loss^{\mr{rec}}_k$, $\Loss^{\mr{rec}}_l$:

\begin{equation}
       \Loss^{GC} = 
       \sum_{k=1}^{N^m} \sum_{\substack{l=1 \\ l \ne k}}^{N^m} \mathcal{I}^{+}_{kl} \log\big(\sigma(GC_{k}^s - GC_{l}^s)\big) \, - \,
       \mathcal{I}^{-}_{kl} \log\big(1-\sigma(GC_{k}^s - GC_{l}^s)\big)
\end{equation}
where $\mathcal{I}^{+}_{kl}= \vOne(\Loss^{\mr{rec}}_k \!>\! \Loss^{\mr{rec}}_l)$, $\mathcal{I}^{-}_{kl} = \vOne(\Loss^{\mr{rec}}_k \!<\! \Loss^{\mr{rec}}_l)$, $\sigma(\cdot)$ is the sigmoid function, and $\Loss^{rec}$ is detailed in Section~\ref{sec: Knowledge Distillation-Guided}.

This loss function enforces consistency between the predicted $GC^s$ values and $\Loss^{\mr{rec}}$ as the \textit{ground truth}, effectively guiding the student model to learn a meaningful ranking of geometric complexities for the masked patches. By comparing all pairs of patches, the loss ensures a robust evaluation of relative complexity within $\InputMasked$. 

\subsubsection{Geometric-Guided Masking}
\label{sec: Geometric-Guided Masking} Patches with a high \gls*{gc} score are typically those that the model struggles to reconstruct accurately (see Fig.~\ref{fig:banner}). This difficulty often stems from their complex geometry, compounded by the absence of color and background information.  
While choosing those patches for masking might seem straightforward, there are two challenges to this approach. 
First, during training, the \gls*{gc} is evaluated by the student for masked patches $\InputMasked$, yet we need to pick candidate patches from the entire set. 
Second, the student's \gls*{gc} estimation can be noisy, making the training unstable. We address both these challenges by instead selecting patches based on the teacher's score (\gls*{gc}\textsuperscript{t}). Thus, at each iteration, the teacher predicts the \gls*{gc} for all patches in $X$, including unmasked ones. Thanks to the momentum update of Fig.~\ref{eq:momentum}, the teacher's predictions are more consistent across different training iterations. 

\subsubsection{Curriculum Mask Selection}
\label{sec:curriculum_patch_selection}
During the initial phases of training, the model may struggle to reconstruct fine details and is often overwhelmed by the complexity of the point cloud structure. 
To mitigate this problem, we follow a curriculum easy-to-hard mask selection strategy by starting from pure random masking at the initial training epoch and gradually increasing the portion of geometric-guided masking until the maximum epoch $e_{\mr{max}}$. Let $A \in [0,1]$ be the maximum ratio of patches that can be selected using \gls*{gc}\textsuperscript{t}. At each epoch $e_t$, we select the $N^{\mr{sel}} = \lfloor e_t/e_{\mr{max}} \times A \times \NumMasked\rfloor$ patches with highest \gls*{gc}\textsuperscript{t} value, and the remaining $\NumMasked - N^{\mr{sel}}$ ones are selected randomly based on a uniform distribution. 

\subsection{Knowledge-Distillation-Guided \gls*{gc}}
\label{sec: Knowledge Distillation-Guided}

Instead of relying exclusively on point geometry, our approach employs a knowledge distillation strategy to also learn from latent features. 
This strategy involves transferring geometric knowledge from a frozen teacher network $\mathrm{F} = (\encoder^f, \decoder^f)$ that processes the full set of patches to the student \gls*{geomask3d}\textsuperscript{s} observing unmasked patches. 
This encourages the student \gls*{geomask3d}\textsuperscript{s} to replicate the feature activations of the knowledge teacher $\mathrm{F}$, indirectly learning from the full structure of data. This unique setup enables the student network to benefit from the global geometric context provided by the teacher network, which is constructed from the complete point cloud. As a result, this process facilitates the learning of robust and meaningful representations, which improve performance on downstream tasks.
The complexity of patches in the feature space is determined by employing the Mean Square Error loss between the output of $\encoder^f$ and the output of $\decoder^s$ before converting back to point space: 
\begin{equation}
    \Loss^{\mr{rec}_f} \, = \, \frac{1}{N^m} \sum_{i=1}^{N^m} 
     \left\|\encoder^{f}\!\big(\InputAll \big)_i - \, \decoder^{s}\!\big([\encoder^{s}\left(\InputVis \right), T_M] \big)_i \right\|^2
\end{equation}
This loss, combined with the Chamfer loss $\Loss^{\mr{rec}_p}$ applied in the point space, serves as the ground-truth loss $\Loss^{\mr{rec}}$ for the prediction of \gls*{gc}:
\begin{equation}
    \Loss^{\mr{rec}} = \Loss^{\mr{rec}_p} + \Loss^{\mr{rec}_f}
    \label{eq:combined}
\end{equation}
The total training loss $\Loss$ is calculated as 
\begin{equation}
    \Loss = \alpha \Loss^{GC} + \beta \Loss^{\mr{rec}_p} + \gamma \Loss^{\mr{rec}_f}
\end{equation}
where $\alpha$, $\beta$, and $\gamma$ are hyper-parameters.

\section{Experiments}

\label{sec:experiments}
Several experiments are carried out to evaluate the proposed method. First, we pretrain both Point-MAE and Point-M2AE networks utilizing our \gls*{geomask3d} approach on the ShapeNet~\citep{chang2015shapenet} training dataset. Moreover, we assess the performance of these pretrained models across a range of standard benchmarks, such as object classification, few-shot learning, and part segmentation. It is important to note that, to maintain a completely fair comparison, we exclusively utilize the encoder of the student network for downstream tasks, ensuring it is identical to the encoder used in the method of interest. 

In our approach, we adopt network configurations consistent with those used in the Point-MAE and Point-M2AE models to guarantee a fully fair comparison, notably using masking ratios of 60\% for Point-MAE and 80\% for Point-M2AE. This involves the technique of dividing point clouds into patches, along with employing the \gls*{knn} algorithm with predetermined parameters for consistent patch uniformity. While our autoencoder architecture, including the configuration of Transformer blocks in both encoder and decoder, generally follows the patterns established in these models, we have uniquely tailored the decoder's design specifically for the \gls*{gc} estimation purposes. 
Moreover, the specifics of our network’s hyperparameters for the pretraining and fine-tuning phases are comprehensively detailed in the Supplementary Materials.

\subsection{Pretraining Setup}
We adopt the ShapeNet dataset~\citep{chang2015shapenet} for the pretraining of our technique, in line with the practices established by Point-MAE and Point-M2AE. This dataset, known for its diverse and extensive collection of 3D models across various categories, provides a robust basis for training and evaluation. It contains 57,448 synthetic 3D shapes of 55 categories. 

After this pretraining phase, we assess the quality of 3D representations produced by our approach through a linear evaluation on the ModelNet40 dataset \citep{wu20153d}. We extract 1,024 points from every 3D model in ModelNet40 and then pass them through our encoder, which remains unchanged during this phase to preserve the learned features. The linear evaluation is performed by a \gls*{svm} fitted on these features. This classification performance is quantified by the accuracy metrics detailed in Table~\ref{tb:3_}. The results clearly indicate that our technique, when applied to Point-MAE and Point-M2AE, enhances the network's performance. 



\subsection{Downstream Tasks}

\mypar{Object Classification on Real-World Dataset.} In self-supervised learning for point clouds, it is crucial to create a model that exhibits strong generalization abilities across various scenarios. The ShapeNet dataset, which is favored for pretraining, contains clean, isolated object models, lacking any intricate scenes or background details. Inspired by this limitation, and building on prior approaches, we put our methods to the test on the ScanObjectNN dataset~\citep{uy2019revisiting}, a more demanding dataset that represents about 15,000 real-world objects across 15 categories. This dataset presents a realistic challenge, with objects that are embedded in cluttered backgrounds, making it an ideal benchmark for assessing our model's robustness and generalization in real-world scenarios.

We carry out tests on three different variants: OBJ-BG, OBJ-ONLY, and PBT50-RS. It is important to note that we do not employ any voting techniques or data augmentation during the testing phase. The outcomes of these experiments can be found in Table~\ref{tb:1}. These results demonstrate that integrating the \gls*{geomask3d} module with Point-MAE and Point-M2AE significantly boosts their object classification accuracy on this dataset. 
These findings underscore our method's effectiveness in complex real-world scenarios.

\mypar{Object Classification on Clean Objects Dataset.} For the task of object classification on the ModelNet40 dataset\citep{wu20153d}, we evaluated our pretrained models using the same protocols and configurations as the Point-MAE approach. ModelNet40, featuring 12,311 pristine 3D CAD models across 40 categories, was divided into a training set of 9,843 models and a testing set of 2,468 models, adhering to established norms. Throughout the training, we employed common data augmentation strategies, including random scaling and shifting. To ensure fair comparisons, the standard voting method~\citep{liu2019relation} was also applied during the testing phase. According to Table~\ref{tb:3}, integrating our  \gls*{geomask3d} module with Point-MAE has yielded a classification accuracy of 94.20\%, which surpasses the performance of the standalone Point-MAE and even the more complex Point-M2AE on this dataset. 

\begin{table}[t!]
\begin{minipage}[t]{0.47\linewidth}
\centering
\caption{\textbf{ Linear evaluation on ModelNet40~\citep{wu20153d} by SVM.}}
\label{tb:3_}
\setlength{\tabcolsep}{5pt} 
\begin{tabular}{l c}
\toprule
Method & SVM \\
\midrule
MAP-VAE~\citesmall{wang2019dynamic} & 88.4 \\
VIP-GAN~\citesmall{guo2021pct} & 90.2 \\
\hline
DGCNN\,+\,Jiasaw~\citesmall{yu2022point} & 90.6 \\
DGCNN\,+\,OcCo~\citesmall{yu2022point} & 90.7 \\
DGCNN\,+\,CrossPoint~\citesmall{yu2022point} & 91.2 \\
Transformer\,+\,OcCo~\citesmall{yu2022point} & 89.6 \\
Point-BERT~\citesmall{yu2022point} & 87.4 \\
\hline
Point-MAE~\citesmall{pang2022masked} & 91.05 \\
\rowcolor{gray!15}
\textBF{Point-MAE\,+\,GM3D} & \textBF{92.30} \\ 
\midrule
Point-M2AE~\citesmall{zhang2022point} & 92.90 \\
\rowcolor{gray!15}
\textBF{Point-M2AE\,+\,GM3D} & \textBF{93.15} \\
\bottomrule
\end{tabular}
\end{minipage}\hfill
\begin{minipage}[t]{0.49\linewidth}
\centering
\caption{\textbf{Ablation study on different maximum hard patch ratios ($A$)}. The highest performance is observed at 50\%, where the OBJ-ONLY score reaches 90.36\%.}
\setlength{\tabcolsep}{5pt}
\begin{tabular}{lcc} 
\toprule
Model & $A$ & \textsc{obj-only} \\
\midrule
Original Point-MAE & 0 & 88.29 \\
Point-MAE+GM3D & 0.4 & 89.67 \\
Point-MAE+GM3D & \textBF{0.5} & \textBF{90.36} \\
Point-MAE+GM3D & 0.7 & 89.84 \\
\bottomrule
\end{tabular}
\label{tb:8}
\end{minipage}
\end{table}

\setlength{\tabcolsep}{-20pt}

\begin{table}[t!]
    \centering
    \begin{minipage}{0.47\textwidth}
        \centering
        \caption{\textbf{Part segmentation on ShapeNetPart~\citep{yi2016scalable}}. mIoU\(_c\) (\%) and mIoU\(_i\) (\%) denote the mean IoU across all part categories and all instances in the dataset, respectively. \selfsup \ represents self-supervised pertaining.
        }
        \label{tb:partseg}
        \setlength{\tabcolsep}{3 pt}
        \begin{tabular}{l c c}
            \toprule
            Method & mIoU\(_c\) & mIoU\(_i\) \\
            \midrule
            PointNet~\citesmall{qi2017pointnet} & 80.39 & 83.70 \\
            PointNet++\citesmall{qi2017pointnet} & 81.85 & 85.10 \\
            DGCNN~\citesmall{wang2019dynamic} & 82.33 & 85.20 \\
            Transformer~\citesmall{yu2022point} & 83.42 & 85.10 \\
            \selfsup Transformer\,+\,OcCo~\citesmall{yu2022point} & 83.42 & 85.10 \\
            \selfsup Point-BERT~\citesmall{yu2022point} & 84.11 & 85.60 \\
            \selfsup I2P-MAE~\citesmall{zhang2023learning} & 85.15 & 86.76 \\
            \selfsup Point-GPT-S~\citesmall{chen2024pointgpt} & 84.10 & 86.2 \\
            \selfsup ACT~\citesmall{dong2022autoencoders} & 84.66 & 86.16 \\
            \midrule
            \selfsup Point-MAE~\citesmall{pang2022masked} & 84.19 & \textBF{86.10} \\
            \rowcolor{gray!15}
            \selfsup\textBF{Point-MAE\,+\,GM3D} & \textBF{84.49} & 86.04 \\
            \midrule
            \selfsup Point-M2AE~\citesmall{zhang2022point} & 84.86 & 86.51 \\
            \rowcolor{gray!15}
            \selfsup\textBF{Point-M2AE\,+\,GM3D} & \textBF{84.91} & \textBF{86.52} \\
            \bottomrule
        \end{tabular}
    \end{minipage}%
    \hfill
    \begin{minipage}{0.48\textwidth}
    \centering
        \caption{\textbf{Linear evaluation on ModelNet40~\citep{wu20153d}.} ‘points’ and ‘Acc’ denote the number of points for training and overall accuracy. \selfsup represents self-supervised pretraining.}
        \label{tb:3}
        \setlength{\tabcolsep}{3pt}
        \begin{tabular}{lcc}
            \toprule
            Method & Points & Acc\,(\%) \\
            \midrule
            PointNet~\citesmall{qi2017pointnet} & 1k & 89.2 \\
            PointNet++~\citesmall{qi2017pointnet} & 1k & 90.5 \\
            \selfsup SO-Net~\citesmall{li2018so} & 5k &  92.5  \\
            DGCNN~\citesmall{wang2019dynamic} & 1k &  92.9 \\
            Point Transformer~\citesmall{zhao2021point} &  & 93.7  \\
            \hline
            Transformer~\citesmall{yu2022point} & 1k & 91.4 \\
            \selfsup Transformer\,+\,OcCo~\citesmall{yu2022point} & 1k & 92.1 \\
            \selfsup Point-BERT~\citesmall{yu2022point} & 1k & 93.2  \\
            \selfsup Point-BERT~\citesmall{yu2022point} & 4k & 93.4  \\
            \selfsup Point-BERT~\citesmall{yu2022point} & 8k & 93.8 \\
            \selfsup Point-M2AE~\citesmall{zhang2022point} & 1k & 94.00 \\
            \selfsup Point-GPT-S~\citesmall{chen2024pointgpt} & 1k & 94.00 \\
            \selfsup ACT~\citesmall{dong2022autoencoders} & 1k & 93.5 \\
            \selfsup I2P-MAE~\citesmall{zhang2023learning} & 1k & 94.1 \\
            \hline
            \selfsup Point-MAE~\citesmall{pang2022masked} & 1k & 93.80 \\
            \rowcolor{gray!15}
            \selfsup\textBF{Point-MAE\,+\,GM3D} & \textBF{1k} & \textBF{94.20}\\ 
            \midrule
        \end{tabular}
    \end{minipage}
\end{table}

\begin{table}[h!]
\centering
\caption{\textbf{Object classification on real-world ScanObjectNN dataset~\citep{uy2019revisiting}.} We evaluate our approach on three variants, among which PB-T50-RS is the hardest setting. Accuracy (\%) for each variant is reported. \selfsup represents self-supervised pretraining.}
\setlength\tabcolsep{5pt}
\label{tb:1}
\begin{tabular}{lccc}
\toprule
Method &  OBJ-BG &  OBJ-ONLY &  PB-T50-RS  \\
\midrule
 PointNet~\citesmall{qi2017pointnet}& 73.3 & 79.2 & 68.0 \\
 SpiderCNN~\citesmall{xu2018spidercnn}& 77.1 & 79.5 & 73.7 \\
 PointNet\(++\)~\citesmall{qi2017pointnet++}& 82.3 & 84.3 & 77.9 \\
 DGCNN~\citesmall{wang2019dynamic}& 82.8& 86.2 & 78.1 \\
 PointCNN~\citesmall{li2018pointcnn}& 86.1 & 85.5 & 78.5 \\
 BGA-DGCNN~\citesmall{uy2019revisiting}& - & - & 79.7 \\
 BGA-PN\(++\)~\citesmall{uy2019revisiting}& - & - & 80.2 \\
 GBNet~\citesmall{qiu2021geometric}& - & - & 80.5 \\
 PRANet~\citesmall{cheng2021net}& - & - & 81.0 \\
\midrule
 Transformer~\citesmall{yu2022point}& 79.86 & 80.55 & 77.24 \\
 \selfsup Transformer-OcCo~\citesmall{yu2022point}& 84.85 & 85.54 & 78.79 \\
 \selfsup Point-BERT~\citesmall{yu2022point}& 87.43 & 88.12 & 83.07 \\
\selfsup I2P-MAE~\citesmall{zhang2023learning}& 94.15 & 91.57 & 90.11 \\
\selfsup Point-GPT-S~\citesmall{chen2024pointgpt}& 91.6 & 90.0 & 86.9 \\ \selfsup ACT~\citesmall{dong2022autoencoders}& 93.29 & 91.91 & 88.21 \\
 \midrule
 \selfsup Point-MAE~\citesmall{pang2022masked} & 90.02 & 88.29 & 85.18 \\
 \rowcolor{gray!15}
 \selfsup \textBF{Point-MAE\,+\,GM3D} & \textBF{93.11} &  \textBF{90.36} & \textBF{88.30} \\
 \midrule
  \selfsup Point-M2AE~\citesmall{zhang2022point} & 91.22 & 88.81 & 86.43 \\
  \rowcolor{gray!15}
 \selfsup \textBF{Point-M2AE\,+\,GM3D} & \textBF{94.14} & \textBF{90.70} & \textBF{87.64} \\
\bottomrule
\end{tabular}
\end{table}

\begin{table}[b!] 
    \centering
    \caption{\textbf{Few-shot classification on ModelNet40}. We report the average accuracy (\%) and standard deviation (\%) of 10 independent experiments. \selfsup represents self-supervised pretraining.}
    \setlength\tabcolsep{5pt} 
    {
    \begin{tabular}{lcccc}
        \toprule
       \multirow[b]{2}{*}{\ Method}  & \multicolumn{2}{c}{5-way} & \multicolumn{2}{c}{10-way} \\
       \cmidrule(l{4pt}r{4pt}){2-3}\cmidrule(l{4pt}r{4pt}){4-5}
        & 10-shot & 20-shot & 10-shot & 20-shot \\
        \midrule
        \ DGCNN~\citesmall{wang2019dynamic} & 91.8\ppm3.7 & 93.4\ppm3.2 & 86.3\ppm6.2 & 90.9\ppm5.1 \\
        \addlinespace[0.3em]
        {\renewcommand{\arraystretch}{0.7} 
        \begin{tabular}[l]{@{}l@{}}\selfsup DGCNN\,+\,OcCo~\citesmall{wang2021unsupervised}\end{tabular}} & 91.9\ppm3.3 & 93.9\ppm3.1 & 86.4\ppm5.4 & 91.3\ppm4.6 \\
        \addlinespace[0.3em]
        \ Transformer~\citesmall{yu2022point} & 87.8\ppm5.2 & 93.3\ppm4.3 & 84.6\ppm5.5 & 89.4\ppm6.3 \\
        \addlinespace[0.3em]
        {\renewcommand{\arraystretch}{0.7}
        \begin{tabular}[l]{@{}l@{}}\selfsup Transformer\,+\,OcCo~\citesmall{yu2022point}\end{tabular}} & 94.0\ppm3.6 & 95.9\ppm2.3 & 89.4\ppm5.1 & 92.4\ppm4.6 \\
        \addlinespace[0.3em]
        \ \selfsup Point-BERT~\citesmall{yu2022point} & 94.6\ppm3.1 & 96.3\ppm2.7 & 91.0\ppm5.4 & 92.7\ppm5.1 \\
        \ \selfsup I2P-MAE~\citesmall{zhang2023learning} & 97.0\ppm1.8 & 98.3\ppm1.3 & 92.6\ppm5.0 & 95.5\ppm3.0 \\
        \ \selfsup Point-GPT-S~\citesmall{chen2024pointgpt} & 96.8\ppm2.0 & 98.6\ppm1.1 & 92.6\ppm4.6 & 95.2\ppm3.4 \\
        \ \selfsup ACT~\citesmall{dong2022autoencoders} & 96.8\ppm2.3 & 98.0\ppm1.4 & 93.3\ppm4.0 & 95.6\ppm2.8 \\
        \ \selfsup Point-M2AE~\citesmall{zhang2022point} & 96.8\ppm1.8 & 98.3\ppm1.4 & 92.3\ppm4.5 & 95.0\ppm3.0 \\
        \midrule
        \ \selfsup Point-MAE & 96.3\ppm2.5 & 97.8\ppm1.8 & 92.6\ppm4.1 & 95.0\ppm3.0 \\
        \addlinespace[0.3em]
        \rowcolor{gray!15}
        \ \selfsup \textBF{Point-MAE\,+\,GM3D} & \textBF{97.0\ppm2.5} & \textBF{98.3\ppm1.3} & \textBF{93.1\ppm4.0} & \textBF{95.2\ppm3.6} \\
        \bottomrule
    \end{tabular}
    } 
    \label{tab:my_label}
\end{table}

\begin{figure}[t!]
\vspace{10pt}
  \centering
  \includegraphics[width=.8\linewidth, height=.6\linewidth]{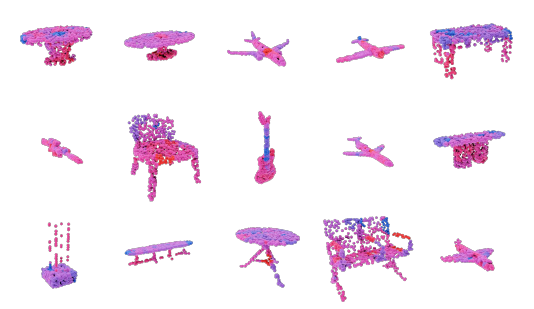} 

  \caption{Visualization of \gls*{gc} values on diverse point clouds from the ShapeNet dataset~\citesmall{chang2015shapenet}.}
  \label{fig:stacked_image}
\end{figure}

\mypar{Few-shot Learning.} Following the protocols of earlier studies~\citep{yu2022point, sharma2020self, wang2021unsupervised}, we conduct few-shot learning experiments on ModelNet40\citep{wu20153d}, using an $n$-way, $m$-shot configuration. Here, $n$ is the number of classes randomly chosen from the dataset, and $m$ is the count of objects randomly selected for each class. The $n\times m$ objects are utilized for training. During the test phase, we randomly sample 20 additional unseen objects from each of the $n$ classes for evaluation.

The results of our few-shot learning experiments are summarized in Fig.~\ref{tab:my_label}. In this highly saturated benchmark, the combination of the \gls*{geomask3d} module exhibits outstanding performance across all tested scenarios. 
It is worth noting that I2P-MAE\citep{zhang2023learning} which \emph{additionally benefits from multiple 2D views} provides only marginal improvements in results. Furthermore, Point-GPT~\citep{chen2024pointgpt} and ACT~\citep{dong2022autoencoders}, despite being \gls*{sota} and complex methods, show only slight improvements compared to each other and other \gls*{sota} methods. Our findings highlight the effectiveness of our method as our Point-MAE+\gls*{geomask3d} model has already achieved higher accuracy than single-scale Point-MAE and multi-scale Point-M2AE. 

\mypar{Part Segmentation.} Our method's capacity for representation learning was assessed using the ShapeNetPart dataset~\citep{yi2016scalable}, which includes 16,881 objects across 16 different categories. In alignment with the approaches taken in prior studies~\citep{qi2017pointnet, qi2017pointnet++, yu2022point}, we sampled 2,048 points from each object to serve as input.

For this highly competitive benchmark, our \gls*{geomask3d} method achieves a slight improvement on both the Point-MAE and Point-M2AE networks, as detailed in Table~\ref{tb:partseg}. Considering that our approach exclusively utilizes 3D information, the observed improvement over methods like I2P-MAE\citep{zhang2023learning} that \emph{supplement 3D with additional 2D data} is reasonable, especially considering the slight enhancements achieved by I2P-MAE. 
Furthermore, Point-GPT \citep{chen2024pointgpt} and ACT \citep{dong2022autoencoders}, despite being \gls*{sota} and complex methods, show only slight improvements over each other and other \gls*{sota} methods. Based on the results of \gls*{sota} methods presented in Table~\ref{tb:partseg}, it is evident that this dataset is highly challenging and competitive. This highlights the effectiveness of our masking strategy in enhancing the understanding of detailed, point-wise 3D patterns.

\subsection{Additional Visualization}
\label{sec:attention map}



\mypar{Geometric Complexity.}
Fig.~\ref{fig:stacked_image} illustrates the \gls*{gc} of randomly selected point clouds from the ShapeNet dataset. This illustration highlights the model's capability to assess \gls*{gc} at the patch level, where the red points denote areas of high \gls*{gc} and blue ones indicate areas of low \gls*{gc}. As mentioned in our methodology section, the model bases the masking process on the predicted \gls*{gc} of the patches. Consequently, patches representing regions with higher \gls*{gc} are preferentially masked. This strategic masking induces the model to focus intensively on intricate point cloud regions containing salient geometric information, thereby enhancing its overall performance in tasks requiring nuanced geometric understanding. 
It is important to note that the provided normalized \gls*{gc} scores are computed relative to the individual patches within each point cloud sample from the ShapeNet dataset. This normalization ensures that the \gls*{gc} scores are a reflection of the variation in complexity within a given sample, enabling the model to internally assess and compare different regions of the same point cloud.

\subsection{Additional Analyses}
\label{sec:abblation2}

\mypar{Pretraining Phase.}
The convergence rates shown in Fig.~\ref{fig:fast_2} clearly highlight the efficiency of our proposed modules. Among the models, the Point-MAE+\gls*{geomask3d} model stands out for its fast convergence, reaching high \gls*{svm} accuracy with fewer epochs compared to the other methods. This quick convergence suggests that the \gls*{geomask3d} module helps the model focus on important features in the data more effectively, speeding up the learning process.

The Point-MAE model, while still effective, achieves a lower accuracy and takes more epochs to get there, indicating that it learns more slowly. On the other hand, the Point-MAE+\gls*{geomask3d}* version (integrating the GM3D method with Point-MAE alongside $\Loss^{\mr{rec}_p}$ and $\Loss^{GC}$) is also effective but doesn’t converge as quickly as the Point-MAE+\gls*{geomask3d}, showing the importance of knowledge distillation alongside \gls*{geomask3d} module.
These findings highlight the practical benefits of adding the \gls*{geomask3d} module to the Point-MAE framework. By helping the model learn faster and more reliably, the \gls*{geomask3d} module not only improves the model's overall performance but also reduces the time needed to achieve high accuracy.



\begin{figure}
  \centering
  \begin{minipage}[b]{0.48\textwidth}
    \includegraphics[width=\linewidth]{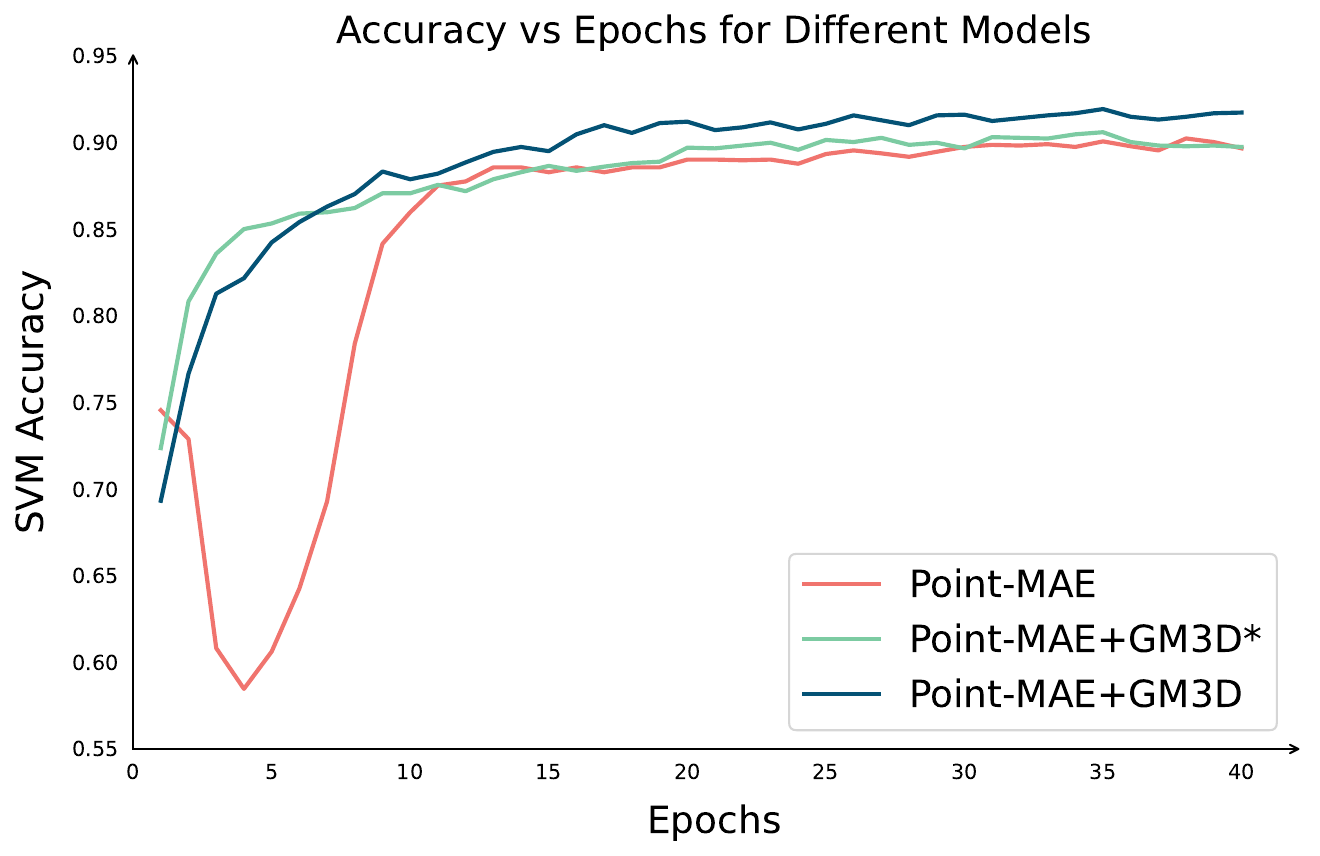} 
    \caption{Comparison of convergence speed during the training phase (Point-MAE).}
    \label{fig:fast_2}
  \end{minipage}\hfill
  \begin{minipage}[b]{0.48\textwidth} 
    \includegraphics[width=\linewidth]
    {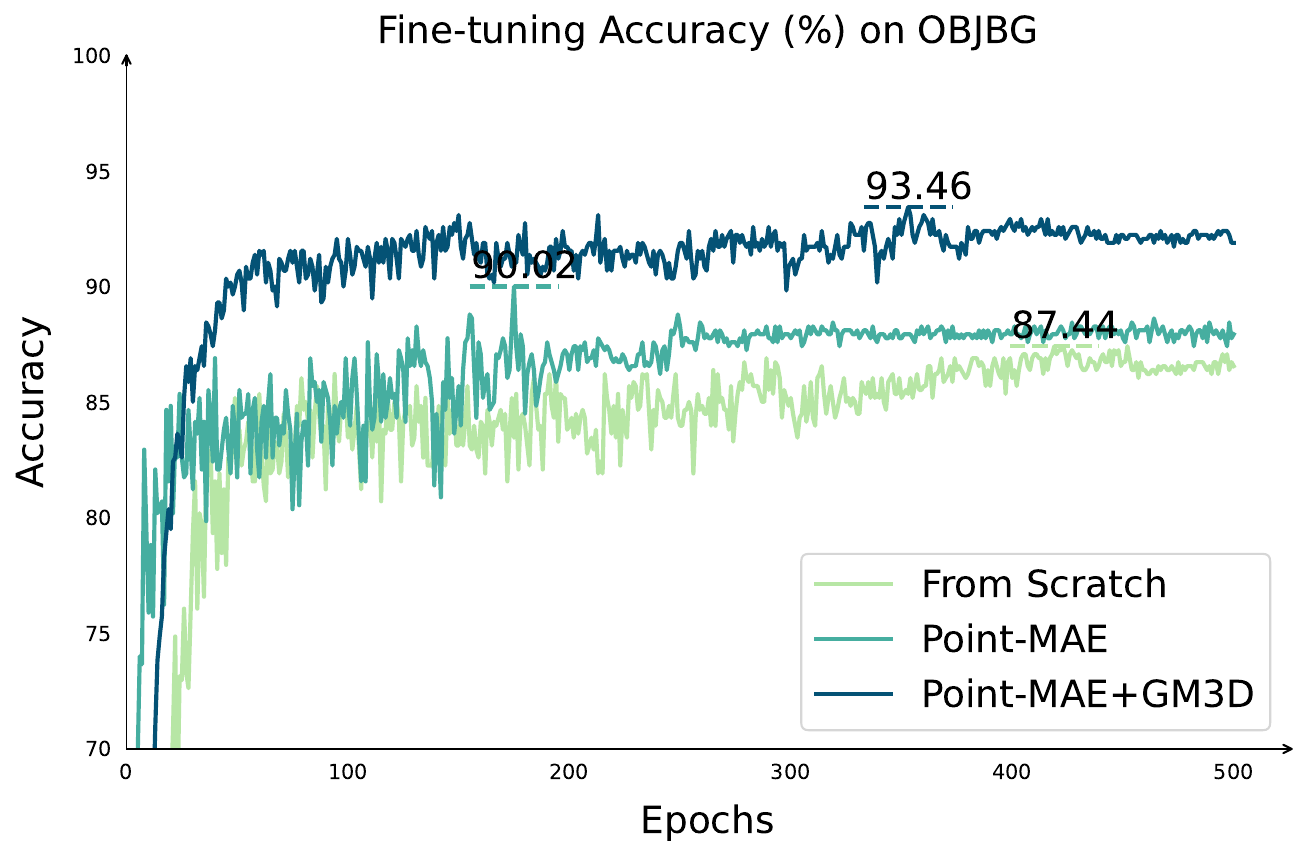}
    \caption{Fine-tuning vs. Training from Scratch on SacnObjectNN (Point-MAE).}
    \label{fig:progress_2}
  \end{minipage}
\end{figure}

\mypar{Fine-tuning Phase.}
Fig.~\ref{fig:progress_2} displays the fine-tuning accuracy on the OBJBG dataset, providing clear evidence of the benefits brought by integrating our \gls*{geomask3d} module with Point-MAE. The results reveal that the Point-MAE+\gls*{geomask3d} model not only achieves the highest accuracy but also maintains this improvement consistently over the course of 400 epochs. This consistent performance highlights the stability and effectiveness of the \gls*{geomask3d} module in guiding the model to learn more relevant features from the data.


\begin{table}[b!]
\centering
\caption{Comparison of Point-MAE, and Point-MAE+GM3D on Pretraining  (\gls*{svm}) and Fine-tuning (OBJ-ONLY) Tasks. `*' stands for our method without $\Loss^{\mr{rec}_f}$.}
\label{tb:loss_ablation}
{
\setlength\tabcolsep{3pt} 
\begin{tabular}{lccc}
\toprule
\ Model & Loss Function & \makecell{SVM\\ModelNet40} & OBJ-ONLY\\
\midrule
Point-MAE & $\Loss^{\mr{rec}_p}$ & 91.05 & 88.29 \\
Point-MAE\,+\,GM3D\textsuperscript{*}~ & $\Loss^{\mr{rec}_p}$\,+\,$\Loss^{GC}$ & \textBF{\textcolor{darkgray}{91.45}} & \textBF{\textcolor{darkgray}{89.50}} \\
Point-MAE\,+\,GM3D & $\Loss^{\mr{rec}_p}$\,+\,$\Loss^{\mr{rec}_f}$\,+\,$\Loss^{GC}$ & \textBF{92.30} & \textBF{90.36} \\
\bottomrule
\end{tabular}
} 
\end{table}

\begin{table}[t]
\centering
\caption{Ablation study on different components of our method based on Point-MAE}
\label{tb:local_ablation}
{
\setlength\tabcolsep{5pt} 
\begin{tabular}{cccccccccc} 
\toprule
& \multicolumn{2}{c}{Input $\shortrightarrow$ $F$} & \multicolumn{2}{c}{$\Loss$} & \multicolumn{2}{c}{$\Loss^{\mr{rec}}$ $\shortrightarrow$ $\Loss^{GC}$} & \multicolumn{1}{c}{${GM3D}^t$} & \multirow[b]{2}{*}{$\Loss^{\mr{rec}_f}$} & \multirow[b]{2}{*}{OBJ-ONLY} \\
\cmidrule(lr){2-3} \cmidrule(lr){4-5} \cmidrule(lr){6-7} 
& $\InputVis$ & $\InputAll$ & $\Loss^{\mr{rec}_f}$ & $\ \Loss^{\mr{rec}_p}$ & 
$\Loss^{\mr{rec}_f}$ & $\Loss^{\mr{rec}_p}$ & $\mu$ \\
\midrule
$a$ & \checkmark & & \checkmark & & \checkmark & & \checkmark & $\encoder^{f} \shortrightarrow \decoder^{s}$ & 89.32\\
$b$ & \checkmark & & \checkmark & \checkmark  & \checkmark & \checkmark & \checkmark & $\encoder^{f} \shortrightarrow \decoder^{s}$ & 90.18\\
$c$ & & \checkmark & \checkmark & \checkmark  & & \checkmark & \checkmark & $\encoder^{f} \shortrightarrow \decoder^{s}$ & 89.67\\
$d$ & & \checkmark & & \checkmark  & & \checkmark & \checkmark & $\encoder^{f} \shortrightarrow \decoder^{s}$ & 89.50\\
$e$ & & \checkmark & \checkmark & \checkmark  & \checkmark & \checkmark & & $\encoder^{f} \shortrightarrow \decoder^{s}$ & 89.33\\
$f$ & & \checkmark & \checkmark & \checkmark  & \checkmark & \checkmark & \checkmark & $\encoder^{f} \shortrightarrow \encoder^{s}$ & 89.15\\
$g$ & & \checkmark & \checkmark & \checkmark  & \checkmark & \checkmark & \checkmark & $\encoder^{f} \shortrightarrow \decoder^{s}$ & \textBF{90.36}\\
\bottomrule
\end{tabular}
} 
\end{table}

\subsection{Ablation Study}

Our ablation study focuses on the incremental improvements offered by our proposed method \gls*{geomask3d} when integrated with the original Point-MAE framework. The original Point-MAE serves as our baseline, using the Chamfer loss for self-supervised learning and setting a performance benchmark on subsequent pretraining and fine-tuning tasks.



\mypar{\gls*{geomask3d}.} Initially, we integrate the \gls*{geomask3d} method with Point-MAE alongside $\Loss^{\mr{rec}_p}$ and $\Loss^{GC}$. As reported in Table~\ref{tb:loss_ablation}, this combination, termed Point-MAE+\gls*{geomask3d}\textsuperscript{*}, shows a clear improvement over the baseline model by achieving higher pretraining SVM evaluation metrics on ModelNet40 and better fine-tuning results on ScanObjectNN (OBJ-ONLY). This supports the idea that a training focus on more geometrically complex patches contributes to improved model generalization.


Building on this structure, we enhance the performance by incorporating knowledge distillation alongside the \gls*{geomask3d} module. The improved model, Point-MAE+\gls*{geomask3d}, which employs three distinct loss functions, not only outperforms the baseline Point-MAE but also shows further improvement over the Point-MAE+\gls*{geomask3d}\textsuperscript{*} approach that utilizes only $\Loss^{\mr{rec}_p}$ and $\Loss^{GC}$. This advancement validates the effectiveness of our knowledge distillation strategy, which focuses on accurate reconstruction while also capturing the complex geometric interrelations in the data. The various impacts of knowledge distillation are further explored in Table~\ref{tb:local_ablation}.




\mypar{Maximum Hard Patch Ratios.} The data presented in Table~\ref{tb:8} offer insights into the ablation study focusing on different hardness ratios, denoted by $A$, within the context of point cloud modeling. It is noteworthy that the inclusion of \gls*{geomask3d} enhances the performance across different $A$ settings when compared to the original model, with the highest performance observed at a 50\% hardness ratio, where the OBJ-ONLY score reaches 90.36\%.

\mypar{Additional Configurations.} In Table~\ref{tb:local_ablation}, which details our ablation study, we investigate the various configurations of our proposed method. The `Input' column pertains to the input utilized by the knowledge-teacher network; it specifies whether complete data $\InputAll$ is provided or only partial data $\InputVis$ are used. The second column, denoted by $\Loss$, encompasses both $\Loss^{\mr{rec}_f}$ and $\Loss^{\mr{rec}_p}$, representing the reconstruction loss functions in two spaces. In the third column, we analyze the impact of the chosen loss functions serving as the ground truth for $\Loss^{GC}$, which is our geometric complexity loss. As can be seen, the performance is enhanced by the geometric complexity guidance, which is informed by the feature-level knowledge distillation (rows c, and g). The subsequent column considers the influence of momentum, a parameter linked to the performance of \gls*{geomask3d}\textsuperscript{t}. In the fifth column, we evaluate the impact of implementing $\Loss^{\mr{rec}_f}$ on the interactions between various components of \gls*{geomask3d}\textsuperscript{s} and $\mathrm{F}$. As evidenced by the results, each setting has been systematically varied to assess its effect on the final performance metric, OBJ-ONLY, demonstrating the significant contributions of each component to the model's learning efficacy.

\section{Conclusion}

We presented a geometrically-informed masked selection strategy for point cloud representation learning. Our \acrfull*{geomask3d} approach leverages a teacher-student model to find complex-to-reconstruct patches in the point cloud, which are more informative for learning robust representations. A knowledge distillation is further proposed to transfer rich geometric information from the teacher to the student, thereby improving the student's reconstruction of masked point clouds. Comprehensive experiments on several datasets and downstream tasks show our method's ability to boost the performance of point cloud learners.

\newpage

\section*{\centering GeoMask3D: Geometrically Informed Mask Selection for Self-Supervised Point Cloud Learning in 3D \\ Supplementary Material}
\vspace{1cm}

\appendix

\section{Implementation Pipeline}

\label{sec:rationale}
We used PyTorch to implement the core functionalities of our approach. The codebase is structured into two primary parts: \emph{main-pretrain} and \emph{main-finetune}.

\mypar{Pretraining Phase (\emph{main-pretrain)}.}
In the \emph{main-pretrain} section, we focus on the initial training phase of our models. In this phase, which is crucial to establish a robust foundation for our models, we first load the ShapeNet dataset and then apply our methods to train the models. \cref{alg:code} summarizes the training steps for \gls*{geomask3d} and includes the pseudocode for the reconstruction of point clouds, generation of geometric complexity, and the distillation of knowledge within the feature space.

\mypar{Finetuning Phase (\emph{main-finetune}).}
Once the pretraining is complete, we proceed to the \emph{main-finetune} part. In this stage, only the encoder of student $\encoder^s$ from the pretrained models is carried forward. The output of this phase is directly responsible for the experimental results presented in our paper.

To ensure complete transparency and reproducibility of our results, we have made all relevant materials publicly available. This includes:
\begin{itemize}
\item The full source code for both \emph{main-pretrain} and \emph{main-finetune} phases.
\item All log files containing the detailed results of our experiments
\item Pretrained models for both pretraining and finetuning stages.
\end{itemize}


\begin{algorithm}[h!]
\caption{Pseudo-Code of \gls*{geomask3d} in a PyTorch-like Style}
\label{alg:code}

\begin{indentedblock}
\textcolor{LimeGreen}{\# teacher inference}  \\
-, $GC^t$ = $\gls*{geomask3d}^t(\InputAll)$ \\
\textcolor{LimeGreen}{\# curriculum patch selection}  \\
$M$ = Mask-Generation($GC^t$, $N^{sel}$)\\
\textcolor{LimeGreen}{\# student forward to compute objectives} \\ 
$X^{rec}$, $f^{rec_s}$, $GC^s$ = $\gls*{geomask3d}^s(\InputVis)$ \\
\textcolor{LimeGreen}{\# knowledge teacher (frozen graph)}  \\
$f^{rec_f}$ = $\mathcal{E}^{f}(\InputAll)$ \\
\textcolor{LimeGreen}{\# compute losses}  \\
\textcolor{LimeGreen}{\# feature-space} \\
$\mathcal{L}^{rec_f} = MSE(f^{rec_s}[M], f^{rec_f}[M])$ \\
\textcolor{LimeGreen}{\# point-space} \\
$\mathcal{L}^{rec_p} = \mathrm{Chamfer}(X^{rec}[M], \InputAll[M])$ \\
\textcolor{LimeGreen}{\# both spaces} \\
$\mathcal{L}^{rec} = \mathcal{L}^{rec_p} + \mathcal{L}^{rec_f}$ \\
\textcolor{LimeGreen}{\# geometric complexity} \\
$\mathcal{L}^{GC} = \mathrm{DRC}(\mathcal{L}^{rec}, GC^{s}, M)$ \\
\textcolor{LimeGreen}{\# final loss} \\
$\mathcal{L} = \mathcal{L}^{rec} +\mathcal{L}^{GC}$ \\
return $\mathcal{L}$
\end{indentedblock}

\end{algorithm}

All these resources can be accessed through our GitHub repository. This repository includes everything needed to understand our code, covering all aspects of the implementations and the reproduction of the results. Moreover, the specifics of our network’s hyperparameters for the pretraining and fine-tuning phases are comprehensively detailed in Table~\ref{tab:hyperparameters}. Additionally, the pre-trained model for the Knowledge Teacher is selected based on the baseline methods, Point-MAE and Point-M2AE.


\begin{table}[b!]
\centering
\caption{Hyperparameter configuration}
\label{tab:hyperparameters}
\small
\renewcommand{\arraystretch}{1.0} 
\setlength{\tabcolsep}{1.7pt}
\begin{tabular}{lc}
\toprule
Config & Value \\ 
\midrule
Optimizer & AdamW \\
Base learning rate & 1e-3 \\
Weight decay & 0.05 \\
Momentum & $\beta_1, \beta_2 = 0.9, 0.95$ \\
$\alpha$ & 1.0 \\
$\beta$ (After epoch 15) & 1000.0 \\
$\gamma$ (After epoch 15) & 10.0 \\
Batch size (Point-MAE+GM3D) & 256 \\
Batch size (Point-M2AE+GM3D) & 128 \\
Learning rate schedule & cosine decay \\
Pre-training epochs & 400 \\
Fine-tuning epochs & 500 \\
Augmentation (pretraining) & random scaling 
and translation \\
Augmentation (finetuning) & random rotation \\
\bottomrule
\end{tabular}

\end{table}

\section{Additional Visualization}
\label{sec:attention map}

\mypar{Geometric Complexity.}
Fig.~\ref{fig:stacked_image} illustrates the \gls*{gc} analysis of randomly selected point clouds from the ShapeNet dataset. This illustration highlights the model's capability to assess \gls*{gc} at the patch level, where the red points denote areas of high \gls*{gc}, while the blue points indicate areas of low \gls*{gc}.

\begin{figure*}[tb]
  \vspace{30pt}
  \centering
  \includegraphics[width=1.0\linewidth, height=1.0\linewidth]{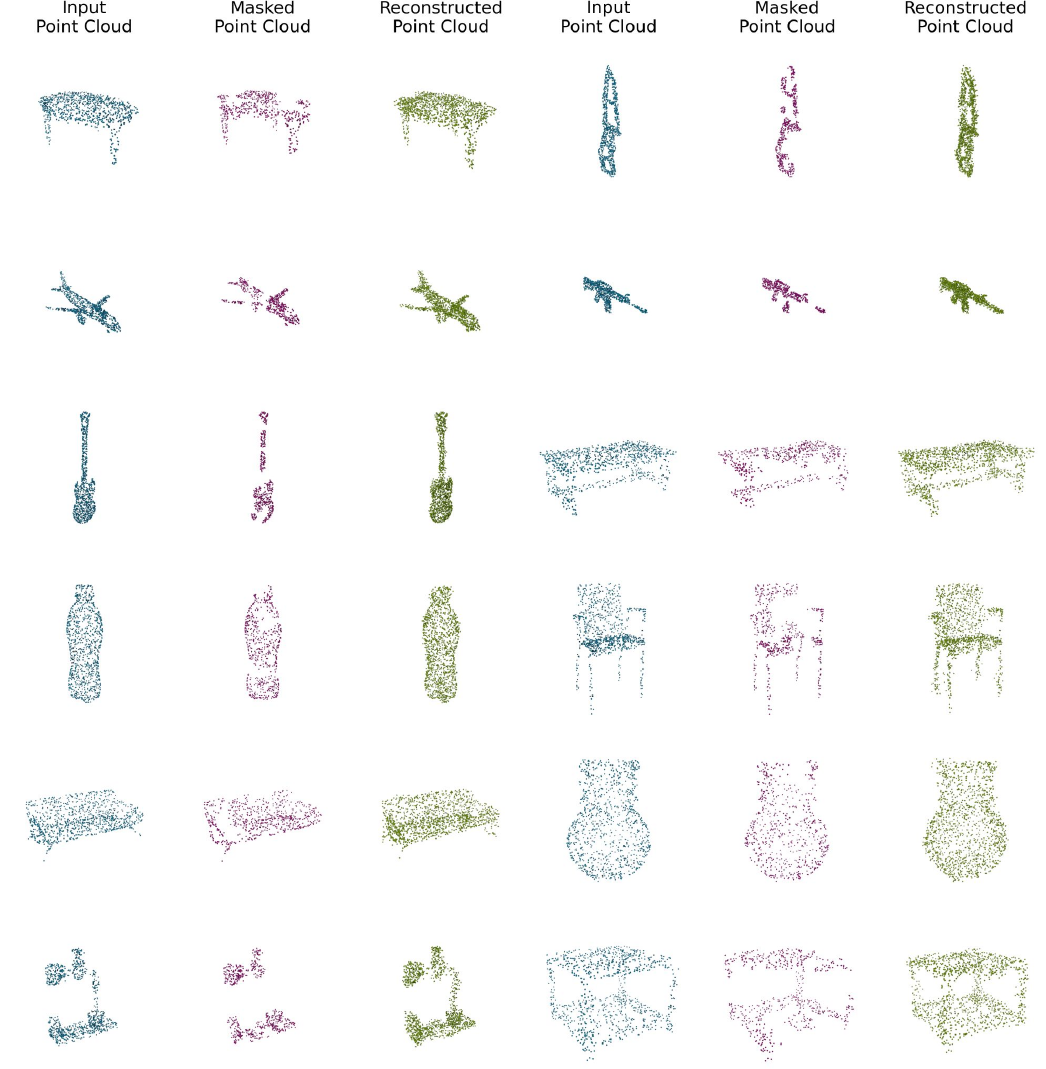}

  \caption{Reconstruction results on the ShapeNet dataset.}
  \label{fig:rec}
\end{figure*}

\mypar{Reconstructed Points.}
To elucidate the capabilities of Masked Autoencoders (MAEs) in processing point cloud data, Fig.~\ref{fig:rec} provides a visual sequence involving the original input, the intermediate masking phase, and the reconstructed output. The first column, titled ``Input Point Cloud'', displays the entirety of the point cloud data, illustrating the initial condition before any processing. The subsequent column, ``Masked Point Cloud'', reveals only the points that remain visible after a portion of the data has been masked. The final column, ``Reconstructed Point Cloud'', demonstrates the model's ability to infer and restore the masked parts of the point cloud. 
The visual comparison in Fig.~\ref{fig:rec} distinctly highlights the high accuracy of the reconstructed points, underscoring the efficacy of our proposed method. It is noteworthy that these visual results were obtained using Point-MAE+\gls*{geomask3d}. For a fair and consistent comparison, the mask ratio used here is like Point-MAE (60\%).


\begin{figure}[tb]
  \centering
  \includegraphics[width=.8\linewidth, height=.4\linewidth]{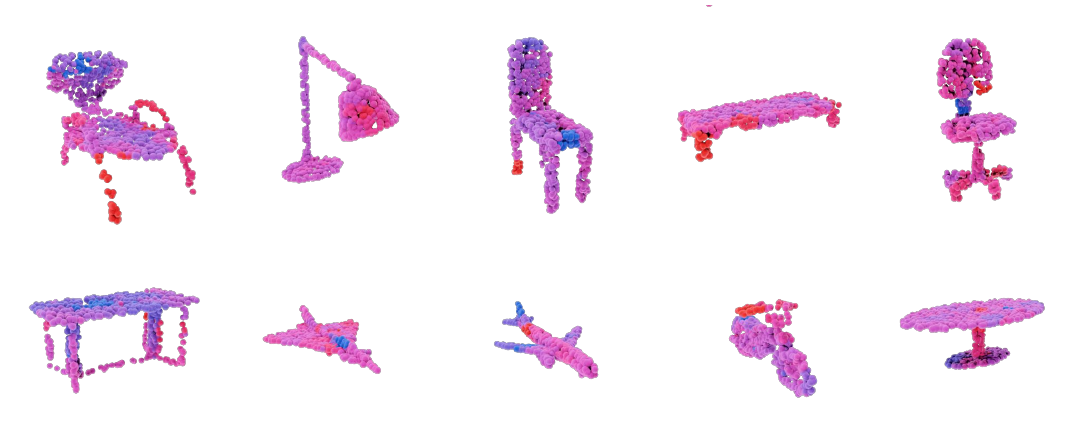} 

  \caption{Visualization of \gls*{gc} values on diverse point clouds from the ShapeNet dataset.}
  \label{fig:stacked_image}
\end{figure}

\section{Additional Analyses}
\mypar{Pretraining Phase.}
In Fig.~\ref{fig:fast_2}, we illustrate the progression of SVM Accuracy throughout the pretraining phase on the ModelNet40 dataset. The red curve represents the Point-M2AE model, while the blue curve denotes the Point-M2AE enhanced with our \gls*{geomask3d} method. It is evident from the graph that the incorporation of \gls*{geomask3d} leads to a substantial improvement in SVM accuracy,  reflecting the model's enhanced classification performance. A key observation is the accelerated convergence rate of the \gls*{geomask3d}-augmented model; it achieves a rapid increase in accuracy within the initial epochs, demonstrating not only the efficacy of \gls*{geomask3d} in facilitating faster learning but also indicating an enhanced ability to generalize from the training data. 

\begin{figure}
  \centering
  \begin{minipage}[b!]{0.48\textwidth}
    \includegraphics[width=\linewidth]{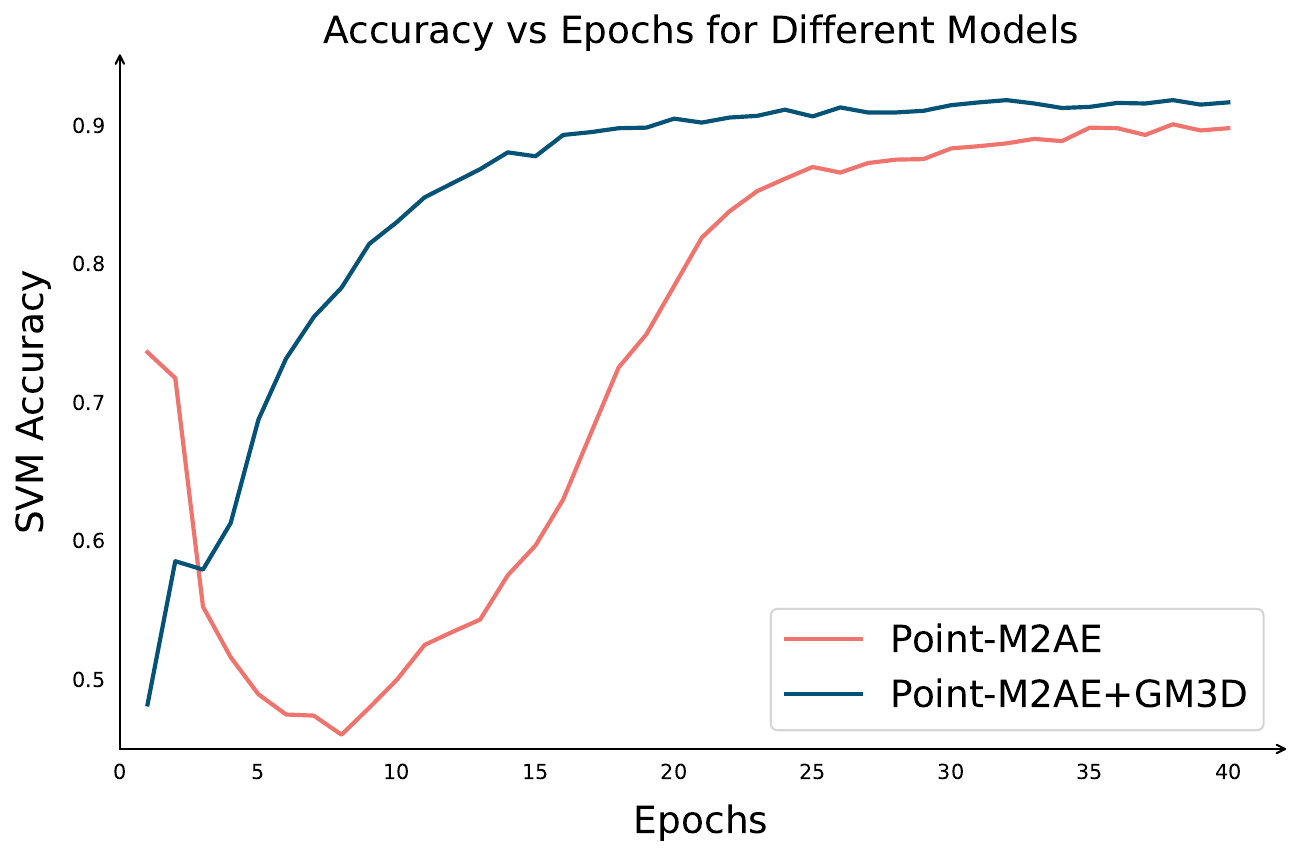} 
    \caption{Comparison of convergence speed during the training phase (Point-M2AE).}
    \label{fig:fast_2}
  \end{minipage}\hfill
  \begin{minipage}[b!]{0.48\textwidth} 
    \includegraphics[width=\linewidth]
    {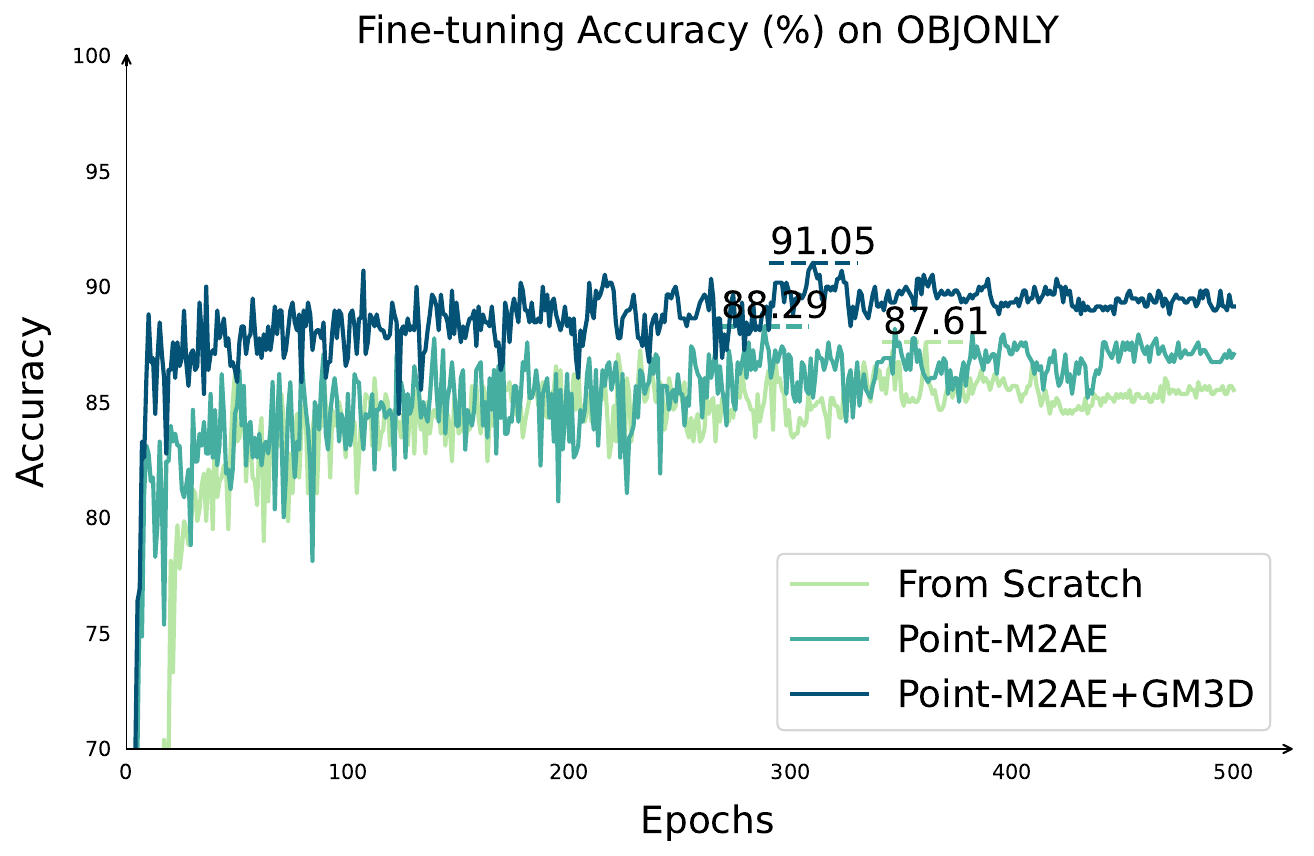}
    \caption{Fine-tuning vs. Training from Scratch on SacnObjectNN (Point-M2AE).}
    \label{fig:fast_44}
  \end{minipage}
\end{figure}

\mypar{Fine-tuning Phase.}
Fig.~\ref{fig:fast_44} displays fine-tuning accuracy on the OBONLY dataset, revealing that the integration of our \gls*{geomask3d} module with Point-M2AE leads to the highest accuracy. Compared to the baseline Point-M2AE and the model trained from scratch, Point-M2AE+\gls*{geomask3d} demonstrates a more significant improvement and exhibits less variability.

\mypar{Integration of GeoMask3D into Point-FEMAE.}
In this section, we present an experimental evaluation of integrating our GeoMask3D approach into the Point-FEMAE \cite{zha2024towards} framework. Point-FEMAE employs two types of masking strategies: global masking and local masking. To enhance its capability, we modified the network by incorporating the geometric complexity decoder $\decoder_{GC}^s$ and replacing the original random global masking strategy with our geometrically guided masking technique.

While Point-FEMAE also utilizes local masking, where Euclidean distances guide the selective masking of tokens related to meaningful parts of the object, this strategy already contributes to capturing geometric structures effectively. 
However, this local masking approach may overlap with the objectives of our GeoMask3D, potentially reducing its distinct impact. Despite this, integrating these two methods provides valuable insights into the effectiveness of geometrically guided masking.

To assess the effectiveness of our approach, we pre-trained Point-FEMAE+GM3D on the ShapeNet dataset and evaluated it on the OBJ-ONLY dataset. The results, presented in Table~\ref{tab:results}, demonstrate that our modification improves performance. For comparison, we reproduced the baseline results of Point-FEMAE on this dataset using the original code and the pre-trained model available in the official repository.


\vspace{+15pt}

\begin{table}[h!]
\centering
\caption{Point-FEMAE with and without GeoMask3D on the OBJ-ONLY dataset}
\setlength\tabcolsep{5pt}
\label{tab:results}
\begin{tabular}{lcc}
\toprule
Method &  OBJ-ONLY  \\
\midrule
 \selfsup Point-FEMAE & 92.08\\
 \rowcolor{gray!15}
 \selfsup \textBF{Point-FEMAE\,+\,GM3D} & \textBF{92.77}\\
\bottomrule
\end{tabular}
\end{table}

\mypar{Impact of Geometric-Guided Masking.}
To further analyze the contribution of geometric-guided masking to the performance improvements observed in our method, we conducted an ablation study to isolate its effect from the \gls{gc} prediction task.


To investigate the isolated impact of \gls{gc} prediction, our method is pre-trained using the Point-MAE backbone with \gls{gc} prediction but without geometric-guided masking. Instead of our masking strategy, we employed random masking. The objective of this experiment was to determine whether \gls{gc} prediction alone contributes to the performance gains or if the synergy with geometric-guided masking is essential. We evaluated the pretrained model on the OBJ-ONLY dataset, as presented in Table~\ref{tab:results_gcm}.


\vspace{+15pt}

\begin{table}[b!]
\centering
\caption{Ablation study on the impact of Geometric-Guided Masking}
\setlength\tabcolsep{5pt}
\label{tab:results_gcm}
\begin{tabular}{lcc}
\toprule
Method &  OBJ-ONLY  \\
\midrule
 \selfsup Point-MAE + GC Prediction + Random Masking & 89.67\\
 \rowcolor{gray!15}
 \selfsup \textBF{Point-MAE + GC Prediction + Geometric-Guided Masking} & \textBF{90.36}\\
\bottomrule
\end{tabular}
\end{table}

As shown in the results, the model trained with both \gls{gc} prediction and geometric-guided masking outperforms the one using \gls{gc} prediction alone. This highlights the necessity of integrating geometric-guided masking with \gls{gc} prediction to achieve optimal performance.

\section{Time Analysis}

In this section, we provide a detailed analysis of the computational resources required by our method compared to the baseline Point-MAE backbone. All experiments were conducted under identical hardware settings using an NVIDIA A6000 GPU with a batch size of 128.

\subsection{Pre-Training}

To evaluate the computational overhead introduced by our approach, we measured the pre-training time per epoch. The results indicate that:

\begin{itemize}
    \item Point-MAE requires approximately 2.8 minutes per epoch.
    \item Our method requires 4.3 minutes per epoch.
\end{itemize}

The additional 1.5 minutes per epoch in our method is not solely due to the inclusion of the knowledge teacher model (which remains frozen during training), but also results from the computational procedures involved in predicting geometric complexity.

\subsection{Downstream Tasks}

As highlighted in the main paper, in downstream tasks, only the encoder of the student model is utilized. Since this encoder is identical to the one used in the baseline methods, our approach maintains computational efficiency during downstream inference. Thus, our method incurs no additional computational burden compared to baseline methods in downstream tasks.

\clearpage 
\bibliographystyle{unsrt}
\bibliography{main.bib} 

\end{document}